\documentclass[review,3p]{elsarticle}
\usepackage{float}
\usepackage{amsmath,amssymb}
\usepackage{graphicx}
\usepackage{booktabs}
\usepackage{multirow}
\usepackage{algorithm}
\usepackage{algorithmic}
\usepackage{hyperref}
\usepackage{xcolor}
\usepackage{subcaption}
\usepackage{placeins}
\usepackage{url}
\usepackage{tikz}
\usetikzlibrary{arrows.meta, positioning, shapes.geometric, calc, fit, backgrounds}

\definecolor{boxInput}{HTML}{E8F1FA}
\definecolor{boxLoss}{HTML}{FFF1D6}
\definecolor{boxOpt}{HTML}{E6F4EA}
\definecolor{boxSqueeze}{HTML}{F5E1F7}
\definecolor{edgeInput}{HTML}{2E5C8A}
\definecolor{edgeLoss}{HTML}{A86B00}
\definecolor{edgeOpt}{HTML}{1E7A3A}
\definecolor{edgeSqueeze}{HTML}{7A2A8A}

\providecommand{\imgOrPlaceholder}[2]{%
  \IfFileExists{#1}{\includegraphics[width=#2]{#1}}%
  {\framebox[#2]{\rule{0pt}{2.2cm}\centering Empty Slot}}%
}

\begin{document}

\begin{frontmatter}

\title{Differentiable Packing of Irregular 3D Objects 
with Adaptive Container Estimation}

\author[iitgn]{Palak Gupta}
\ead{palak.gupta@iitgn.ac.in}
\author[iitgn]{Shanmuganathan Raman}
\ead{shanmuga@iitgn.ac.in}
\address[iitgn]{Indian Institute of Technology Gandhinagar, India}

\begin{abstract}
Most existing approaches either fix the container in advance or optimize only a single container dimension through an outer search loop, leaving the remaining dimensions as a manual tuning problem. We present a differentiable packing framework that jointly optimizes all $6N$ object pose parameters and all three container side lengths inside a single gradient-based loop. The formulation combines six physics-inspired, differentiable loss terms computed directly on triangle meshes through axis-aligned bounding-box proxies. An adaptive squeezing mechanism periodically tightens the container whenever the overlap loss falls below a pair-count-scaled threshold, producing a large initial drop in container volume, followed by small refinements. All pairwise computations are written in tensor-broadcasting form, giving a 3.4 to 54 times speedup over a reference loop-based implementation. The pipeline is implemented in Python and PyTorch, with no physics engine, FFT library, or convex decomposition. On multiple object categories, the method produces containers that are 11 to 32 percent smaller than time-matched DBLF and simulated-annealing baselines at $N{=}100$, while running in under 4 minutes per instance on a single consumer GPU.

\end{abstract}


\end{frontmatter}
\begin{figure*}[!htbp]
\centering
\includegraphics[width=0.72\linewidth]{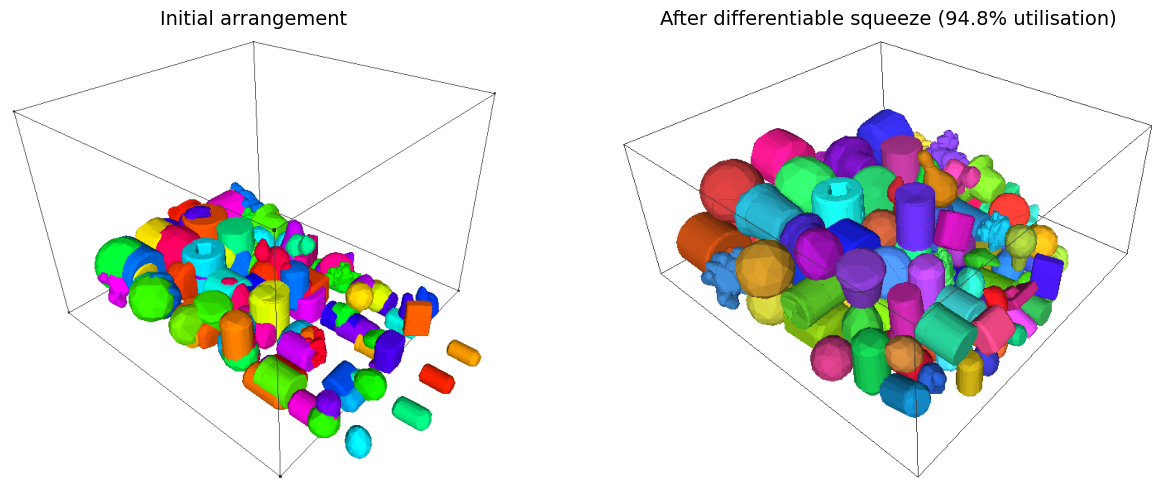}
\caption{Our differentiable framework jointly optimizes object poses and all 
three container dimensions in a single gradient-based loop. \textbf{Left:} 
initial grid arrangement at $N{=}100$ kitchen objects inside the loose 
estimated container. \textbf{Right:} converged arrangement after pose 
optimization and adaptive container squeezing, achieving $94.8\%$ container 
utilization with zero true mesh--mesh intersection.}
\label{fig:teaser}
\end{figure*}

\section{Introduction}
\label{sec:intro}

The 3D packing problem asks to arrange a set of objects inside a container without overlap, typically as tightly as possible. For axis-aligned cuboids, the problem has been studied for decades~\cite{bortfeldt2013constraints,zhao2016comparative}. Real-world objects, however, are rarely cuboids: mechanical parts, kitchen utensils, 3D-printed assemblies, and consumer products are geometrically irregular and often concave. Discrete placement rules leave large voids on such inputs, and methods that replace irregular meshes with their axis-aligned bounding boxes as a permanent input proxy are bounded by the looseness of that replacement. Our method evaluates AABBs only as a per-iteration overlap surrogate computed from the transformed mesh vertices; the triangle mesh itself drives every pose update.

Two variants dominate the literature. The \emph{bin packing problem} (BPP) asks how many objects fit inside a container of given dimensions. The \emph{open dimension problem} (ODP) asks to minimize one or more container dimensions while still accommodating every object. Existing approaches address these variants through constructive heuristics~\cite{wang2010two,wang2022dense}, metaheuristic search~\cite{tole2023fgs}, rigid-body simulation~\cite{zhuang2024dynamics}, continuous optimisation~\cite{ma2018packing,romanova2018packing}, and reinforcement learning~\cite{zhao2023learning}.

Irregular packing arises particularly in additive manufacturing, 
where build-plate utilisation directly affects machine 
time~\cite{araujo2019irregular, calabrese2022nesting}.
Leao et al.~\cite{leao2020irregular} provide a comprehensive 
review of mathematical models for irregular packing problems.
Hopper and Turton~\cite{hopper2001review} survey 2D strip 
packing, which shares structural similarities with the 3D case.

\paragraph{The gap} Existing ODP methods share two structural limitations. \textbf{(i)} Container optimisation is one-dimensional: Zhuang et al.~\cite{zhuang2024dynamics} optimise only the height; Ma et al.~\cite{ma2018packing} run a binary search on a single dimension, invoking a continuous optimiser at every step. The other two dimensions are fixed in advance. \textbf{(ii)} The strongest recent results~\cite{zhuang2024dynamics,cui2023dense} are built on C\texttt{++} and CUDA pipelines combining NVIDIA PhysX with cuFFT, which are not easy to reproduce, extend, or integrate into a PyTorch or JAX workflow.

\paragraph{Our approach} We present a differentiable packing framework that optimizes the full $6N$-dimensional pose vector jointly with all three container side lengths in a single continuous loop. The method operates in two modes sharing the same inner engine: a \emph{fixed-container} mode for the BPP, and a \emph{minimum container estimation} mode that progressively shrinks the container via an adaptive squeezing mechanism. Objects are represented directly as triangle meshes, with no voxelization, signed distance fields, or convex decomposition. Six physics-inspired differentiable loss terms drive the optimization, and every pairwise interaction is expressed as a broadcast tensor operation, so the forward pass contains no Python-level pair loops. Figure~\ref{fig:architecture} summarises the pipeline.

\paragraph{Contributions}
\begin{enumerate}
\item A differentiable formulation of 3D irregular packing with six physics-motivated loss terms (overlap, boundary, support-aware gravity, contact, cohesion, centripetal), each with a meaningful gradient through the rotation and translation parameters.
\item An adaptive container squeezing mechanism with a pair-count-scaled overlap threshold that makes the same procedure work reliably from $N{=}10$ to $N{>}100$ without per-$N$ tuning.
\item A fully vectorized PyTorch implementation that replaces the $\binom{N}{2}$ Python loop with tensor broadcasting, yielding a 3.4 to 54 times speedup and allowing a 100-object instance to finish in about three minutes on a single consumer GPU.
\item A systematic evaluation on multiple object categories against three time-matched baselines. At $N{=}100$, our method produces containers 11 to 32 percent smaller than the best of DBLF, simulated annealing, and BLF+SA.
\end{enumerate}

\paragraph{Reproducibility} The entire framework depends only on PyTorch, NumPy, SciPy, and \texttt{trimesh}. No compiled extensions, custom CUDA kernels, physics engine bindings, or FFT libraries are used. All hyperparameters are reported in Section~\ref{sec:experiments}, and all results are generated with five fixed random seeds.

\FloatBarrier
\section{Related Work}
\label{sec:related}

\paragraph{Constructive heuristics}
Wang et al.~\cite{wang2010two} introduced Deepest-Bottom-Left-Fill (DBLF), a sequential greedy placement rule. Wang and Hauser~\cite{wang2022dense} extended DBLF with a heightmap-minimizing step. The same authors address robotic 
packing with nondeterministic item arrival~\cite{wang2019robot}, 
a related but distinct online setting. Crainic et al.~\cite{crainic2008extreme} generalized such rules through an extreme-point framework. DBLF-family methods are fast and yield zero-overlap arrangements by construction, but their greedy nature makes them dependent on placement order, and they cannot revisit earlier decisions. Garey  et al.~\cite{garey1979computers} established the 
NP-hardness of bin packing, which motivates the use of 
heuristic and continuous optimisation approaches.

\paragraph{Metaheuristic search}
Simulated annealing~\cite{tole2023fgs} and genetic algorithms~\cite{gehring2002genetic} have been applied by encoding a candidate solution as the object sequence passed to a DBLF-like inner placer. These methods escape strict greediness but remain bottlenecked by the inner discrete placer and by the curse of dimensionality as $N$ grows. 

\paragraph{Physics-based simulation}
Zhuang et al.~\cite{zhuang2024dynamics} combine DBLF initialization with FFT-based collision detection~\cite{cui2023dense} and rigid-body dynamics in NVIDIA PhysX. Horizontal shaking drives objects into compact arrangements. The pipeline is built on C\texttt{++}, CUDA, PhysX, and cuFFT, and optimizes only the container height.

\paragraph{Continuous optimisation}
Ma et al.~\cite{ma2018packing} use continuous optimization of positions and orientations followed by combinatorial object-swapping, running a binary search over the height dimension and reporting 28 to 58 minutes for 60 to 130 objects. Stoyan et al.~\cite{stoyan2016cutting} develop quasi-phi-functions 
for continuous rotations, on which Romanova et al.~\cite{romanova2018packing} 
build for concave polyhedra but scaling to only about 10 objects.

\paragraph{Spectral and learning-based methods}
Cui et al.~\cite{cui2023dense} voxelise objects and use 
FFT-based spectral correlation, restricting rotations to a 
discrete set and assuming a fixed container. 
Learning-based approaches~\cite{zhao2023learning, zhao2021online} 
train RL agents for online BPP with a fixed container; 
Hu et al.~\cite{hu2017solving} apply DRL to cuboid packing. 
All three settings assume a fixed container and are not 
directly comparable to our offline container estimation problem.

\paragraph{Geometry processing and fabrication-aware packing}
Attene~\cite{attene2015shapes} introduces a 3D packing 
algorithm that splits objects into parts for tight 
box-packing with minimum container volume, addressing 
the same NP-hard problem class as our work from a 
fabrication-oriented perspective. 
Cao et al.~\cite{cao2021stacking} address constrained 
3D stacking for additive manufacturing, proposing a voxel-based layout 
method for DLP printing that highlights the importance 
of physical constraints in packing. 
Krs et al.~\cite{krs2021pico} present a procedural 
iterative constrained optimiser for geometric modelling 
that combines gradient-based optimisation 
with geometric constraints, closely aligned 
with our differentiable formulation. 

\paragraph{Positioning} Our method differs in three ways: all three container dimensions are jointly optimized; optimization runs directly on triangle meshes via per-epoch AABBs, without voxelization or FFT grid; and the pipeline runs in pure Python and PyTorch.

\FloatBarrier
\section{Method}
\label{sec:method}

\subsection{Overview}
\label{sec:overview}

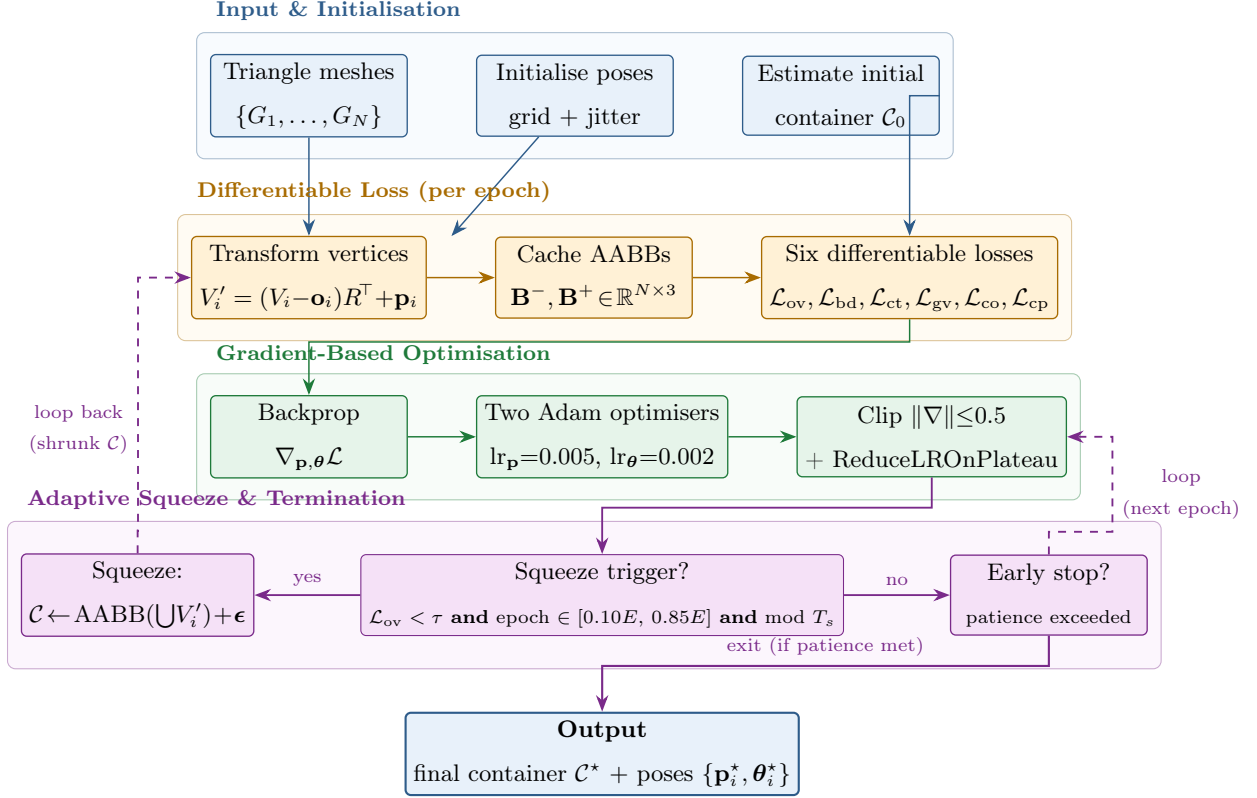
\begin{figure*}[!ht]
\centering
\begin{tikzpicture}[
    font=\small,
    node distance=6mm and 9mm,
    >={Stealth[length=2.2mm, width=1.8mm]},
    box/.style={draw, rounded corners=2pt, align=center,
        minimum width=26mm, minimum height=9mm, inner sep=3pt, line width=0.5pt},
    input/.style={box, fill=boxInput,   draw=edgeInput},
    loss/.style={box,  fill=boxLoss,    draw=edgeLoss},
    opt/.style={box,   fill=boxOpt,     draw=edgeOpt},
    squeeze/.style={box, fill=boxSqueeze, draw=edgeSqueeze},
    deci/.style={draw=edgeSqueeze, fill=boxSqueeze, rounded corners=2pt,
        align=center, inner sep=3pt, minimum width=28mm},
    flow/.style={-{Stealth[length=2.2mm]}, line width=0.55pt, draw=black!70},
    flowblue/.style={flow, draw=edgeInput},
    floworange/.style={flow, draw=edgeLoss},
    flowgreen/.style={flow, draw=edgeOpt},
    flowpurple/.style={flow, draw=edgeSqueeze, line width=0.7pt},
    grouplbl/.style={font=\footnotesize\bfseries, text=black!70}
]
\node[input] (meshes)    {Triangle meshes\\$\{G_1,\ldots,G_N\}$};
\node[input, right=of meshes]  (initpose)  {Initialise poses\\ grid $+$ jitter};
\node[input, right=of initpose] (initcont) {Estimate initial\\container $\mathcal{C}_0$};

\node[loss, below=13mm of meshes]
    (transform) {Transform vertices\\$V_i' = (V_i{-}\mathbf{o}_i)R^{\!\top}{+}\mathbf{p}_i$};
\node[loss, right=of transform]
    (aabb) {Cache AABBs\\$\mathbf{B}^{-}, \mathbf{B}^{+}\!\in\!\mathbb{R}^{N\times 3}$};
\node[loss, right=of aabb]
    (losses) {Six differentiable losses\\$\mathcal{L}_\text{ov}, \mathcal{L}_\text{bd}, \mathcal{L}_\text{ct}, \mathcal{L}_\text{gv}, \mathcal{L}_\text{co}, \mathcal{L}_\text{cp}$};

\node[opt, below=10mm of transform]
    (backprop) {Backprop\\$\nabla_{\mathbf{p},\boldsymbol{\theta}}\mathcal{L}$};
\node[opt, right=of backprop]
    (adam) {Two Adam optimisers\\$\text{lr}_{\mathbf{p}}{=}0.005$, $\text{lr}_{\boldsymbol{\theta}}{=}0.002$};
\node[opt, right=of adam]
    (sched) {Clip $\|\nabla\|{\le}0.5$\\+ ReduceLROnPlateau};

\node[deci, below=10mm of adam]
   (trigger) {Squeeze trigger?\\
              \scriptsize $\mathcal{L}_{\text{ov}}<\tau$ {\bfseries and}
              epoch $\in [0.10E,\,0.85E]$ {\bfseries and} mod $T_s$};

\node[squeeze, left=14mm of trigger]
   (squeeze) {Squeeze:\\$\mathcal{C}\!\leftarrow\!\mathrm{AABB}(\bigcup\! V_i')\!+\!\boldsymbol{\epsilon}$};

\node[squeeze, right=14mm of trigger]
   (earlystop) {Early stop?\\\scriptsize patience exceeded};

\node[input, below=10mm of trigger, draw=edgeInput, line width=0.8pt]
   (output) {\textbf{Output}\\final container $\mathcal{C}^\star$ $+$ poses $\{\mathbf{p}_i^\star,\boldsymbol{\theta}_i^\star\}$};

\draw[flowblue] (meshes)   -- (transform);
\draw[flowblue] (initpose) -- ($(transform.north)!0.5!(aabb.north)$);
\draw[flowblue] (initcont) -| (losses.north);
\draw[floworange] (transform) -- (aabb);
\draw[floworange] (aabb)      -- (losses);
\draw[flowgreen] (losses.south) -- ++(0,-3mm) -| (backprop.north);
\draw[flowgreen] (backprop) -- (adam);
\draw[flowgreen] (adam)     -- (sched);
\draw[flowpurple] (sched.south) -- ++(0,-4mm) -| (trigger.north);
\draw[flowpurple] (trigger.west) -- node[above, font=\scriptsize, text=edgeSqueeze]{yes} (squeeze.east);
\draw[flowpurple] (trigger.east) -- node[above, font=\scriptsize, text=edgeSqueeze]{no}  (earlystop.west);
\draw[flowpurple, dashed]
    (squeeze.north) -- ++(0,3mm) -| ([xshift=-7mm]transform.west)
    node[pos=0.7, left, font=\scriptsize, text=edgeSqueeze, align=center]{loop back\\(shrunk $\mathcal{C}$)}
    -- ([xshift=-6mm]transform.west |- transform.west) -- (transform.west);
\draw[flowpurple, dashed]
    (earlystop.north) -- ++(0,3mm) -| ([xshift=6mm]sched.east)
    node[pos=0.7, right, font=\scriptsize, text=edgeSqueeze, align=center]{loop\\(next epoch)}
    -- ([xshift=6mm]sched.east |- sched.east) -- (sched.east);
\draw[flowpurple, line width=0.8pt]
    (earlystop.south) -- ++(0,-4mm) -|
    node[pos=0.25, above, font=\scriptsize, text=edgeSqueeze]{exit (if patience met)}
    (output.north);

\begin{pgfonlayer}{background}
  \node[fit=(meshes)(initcont), draw=edgeInput!40, rounded corners=3pt,
        fill=boxInput!30, inner xsep=5pt, inner ysep=8pt] (g1) {};
  \node[fit=(transform)(losses), draw=edgeLoss!40, rounded corners=3pt,
        fill=boxLoss!30, inner xsep=5pt, inner ysep=8pt] (g2) {};
  \node[fit=(backprop)(sched), draw=edgeOpt!40, rounded corners=3pt,
        fill=boxOpt!30, inner xsep=5pt, inner ysep=8pt] (g3) {};
  \node[fit=(squeeze)(trigger)(earlystop), draw=edgeSqueeze!40, rounded corners=3pt,
        fill=boxSqueeze!30, inner xsep=5pt, inner ysep=12pt] (g4) {};
\end{pgfonlayer}

\node[grouplbl, anchor=south west] at ([xshift=4pt,yshift=1pt]g1.north west) {\textcolor{edgeInput}{Input \& Initialisation}};
\node[grouplbl, anchor=south west] at ([xshift=4pt,yshift=1pt]g2.north west) {\textcolor{edgeLoss}{Differentiable Loss (per epoch)}};
\node[grouplbl, anchor=south west] at ([xshift=4pt,yshift=0pt]g3.north west) {\textcolor{edgeOpt}{Gradient-Based Optimisation}};
\node[grouplbl, anchor=south west] at ([xshift=4pt,yshift=1pt]g4.north west) {\textcolor{edgeSqueeze}{Adaptive Squeeze \& Termination}};
\end{tikzpicture}
\caption{Pipeline architecture. Meshes and the initial container enter the loss block (amber), which caches per-object AABBs from the transformed vertices and evaluates six differentiable loss terms. Gradients flow into the optimization block (green). The squeeze block (purple) tightens the container when the overlap loss is below the pair-count-scaled threshold $\tau$. Early stopping exits the loop when no improvement is observed for the patience window.}
\label{fig:architecture}
\end{figure*}

Given $N$ triangle-mesh objects, we seek a position $\mathbf{p}_i \in \mathbb{R}^3$ and Euler-angle rotation $\boldsymbol{\theta}_i \in \mathbb{R}^3$ for every object such that (i) no two objects overlap, (ii) all objects lie inside a rectangular container whose volume is minimised, and (iii) each object rests on the ground plane or another object. The framework (Figure~\ref{fig:architecture}) has four blocks: an input block that loads meshes and estimates an initial container; a loss block that computes six differentiable loss terms from cached AABBs; an optimization block that steps two Adam optimizers over the $6N$ pose parameters; and a squeeze block that monitors the overlap loss and updates the container when triggered. Algorithm~\ref{alg:pipeline} gives the pseudo-code.

\begin{algorithm}[t]
\caption{Differentiable packing with adaptive squeezing.}
\label{alg:pipeline}
\begin{algorithmic}[1]
\REQUIRE meshes $\{G_1,\ldots,G_N\}$, initial container $\mathcal{C}_0$, max epochs $E_{\max}$
\ENSURE final container $\mathcal{C}^\star$, poses $\{(\mathbf{p}_i^\star,\boldsymbol{\theta}_i^\star)\}$
\STATE initialise $\{\mathbf{p}_i\}$ on a compact grid with Gaussian jitter; set $\{\boldsymbol{\theta}_i\}\gets\mathbf{0}$
\STATE set squeeze threshold $\tau \gets \max(5.0,\ 0.005 \cdot N(N-1)/2)$
\STATE set squeeze window $[\,E_{\text{lo}},E_{\text{hi}}\,]\gets[\lfloor 0.10 E_{\max}\rfloor,\lfloor 0.85 E_{\max}\rfloor]$
\STATE set squeeze period $T_s\gets\max(40,\lfloor E_{\max}/20\rfloor)$
\FOR{$e=1,\ldots,E_{\max}$}
  \STATE transform vertices: $V_i' \gets (V_i{-}\mathbf{o}_i)R(\boldsymbol{\theta}_i)^\top+\mathbf{p}_i$
  \STATE cache AABBs; evaluate $\mathcal{L}_{\text{total}}$ via vectorised broadcasts
  \STATE backpropagate; clip $\|\nabla\|_2\le 0.5$; step both Adam optimisers
  \IF{$\mathcal{L}_{\text{ov}}<\tau$ \AND $e\in[E_{\text{lo}},E_{\text{hi}}]$ \AND $e\bmod T_s = 0$}
      \STATE $\mathcal{C}\gets \mathrm{AABB}(\bigcup_i V_i') + \boldsymbol{\epsilon}$ \quad\COMMENT{squeeze}
  \ENDIF
  \IF{early-stop patience exceeded} \STATE \textbf{break} \ENDIF
\ENDFOR
\RETURN $\mathcal{C}^\star, \{(\mathbf{p}_i^\star,\boldsymbol{\theta}_i^\star)\}$
\end{algorithmic}
\end{algorithm}

\subsection{Object Representation and Transformation}
\label{sec:representation}

Each mesh $G_i = (V_i, F_i)$ is re-centered at its object-local bounding-box centroid $\mathbf{o}_i = \tfrac{1}{2}(\min V_i + \max V_i)$. For Euler angles $\boldsymbol{\theta}_i = (\theta_x,\theta_y,\theta_z)$ in ZYX convention, the rotation matrix is
\begin{equation}
R(\boldsymbol{\theta}_i) = R_z(\theta_z)\,R_y(\theta_y)\,R_x(\theta_x) \in SO(3),
\end{equation}
and the transformed vertices are
\begin{equation}
V_i' = (V_i - \mathbf{o}_i)\,R(\boldsymbol{\theta}_i)^\top + \mathbf{p}_i.
\end{equation}
The instantaneous AABB is $\mathbf{b}_i^- = \min V_i'$, $\mathbf{b}_i^+ = \max V_i'$, recomputed every iteration from the transformed vertices. Rotating a precomputed object-local AABB would yield a strict over-approximation of the rotated object's AABB and forfeit the geometric information that makes tight irregular packing possible.

Each object contributes 6 learnable parameters; the full search space has dimension $6N$. Objects are sorted by volume in decreasing order and placed on a coarse grid of side $G = \lceil N^{1/3}\rceil - 1$ with Gaussian jitter ($\sigma{=}0.1$) in the $x$ and $z$ axes; $y$ is set to the ground plane. Sorting by volume empirically reduces the number of epochs spent unwedging a large object from between smaller neighbors. Rotations are initialized to $\mathbf{0}$.

\subsection{Physics-Inspired Loss Functions}
\label{sec:losses}

The total loss is a weighted sum of six terms:
\begin{equation}
\mathcal{L}_{\text{total}} = \mathcal{L}_{\text{ov}} + \mathcal{L}_{\text{bd}} + \tfrac{1}{\sqrt{N}}\,\mathcal{L}_{\text{ct}} + \mathcal{L}_{\text{gv}} + \mathcal{L}_{\text{co}} + \mathcal{L}_{\text{cp}}.
\label{eq:total_loss}
\end{equation}
The $1/\sqrt{N}$ prefactor on the contact term prevents a quadratic-in-$N$ attractive force from overwhelming the overlap term at large $N$.

\paragraph{Overlap Loss} Per-axis overlap between two AABBs is
$o_{ij}^k = \max(0,\min(b_{i,k}^+,b_{j,k}^+) - \max(b_{i,k}^-,b_{j,k}^-))$ for $k\in\{x,y,z\}$, and
\begin{equation}
\mathcal{L}_{\text{ov}} = w_{\text{ov}}\sum_{i<j}\bigl(\prod_{k} o_{ij}^k\bigr)\mathbf{1}[\forall k:\,o_{ij}^k > \epsilon],
\end{equation}
with $w_{\text{ov}} = 80\ln(N{+}1)$ and $\epsilon = 10^{-6}$. The log scaling keeps the term dominant during the early phase of optimization, independent of $N$.

\paragraph{Boundary Loss} Objects must stay inside the container $[\mathbf{c}^-,\mathbf{c}^+]$, with an amplified ground term:
\begin{equation}
\begin{aligned}
\mathcal{L}_{\text{bd}} = w_{\text{bd}}\sum_i\Bigl[
  &100\sum_k\bigl(\max(0,c_k^-{-}b_{i,k}^-)^2 + \max(0,b_{i,k}^+{-}c_k^+)^2\bigr) \\
  &+ 1000\,\max(0,-b_{i,y}^-)^2\Bigr].
\end{aligned}
\end{equation}

\paragraph{Support-Aware Gravity Loss} A naive penalty on $b_{i,y}^-$ would prevent stacking. Instead, the effective supporting height is computed from objects beneath whose horizontal footprints overlap by at least $\eta$:
\begin{equation}
\mathcal{S}_i = \{\,j : o_{ij}^x > \eta,\ o_{ij}^z > \eta,\ b_{j,y}^+ \le b_{i,y}^-\,\},\quad
g_i = \max(0,\max_{j\in\mathcal{S}_i} b_{j,y}^+).
\end{equation}
With floating gap $\gamma_i = \max(0,b_{i,y}^- - g_i)$,
\begin{equation}
\mathcal{L}_{\text{gv}} = \sum_i (20\gamma_i^2 + 5\gamma_i)\mathbf{1}[\gamma_i>10^{-3}] + 2000\sum_i \max(0,-b_{i,y}^-)^2.
\end{equation}
This produces emergent multi-layer stacking without any explicit heuristic.

\paragraph{Contact, Cohesion, Centripetal} With per-axis gap $d_{ij}^k$,
\begin{equation}
\mathcal{L}_{\text{ct}} = w_{\text{ct}}\sum_{i<j}\sum_k (d_{ij}^k)^2 \mathbf{1}[\lVert d_{ij}\rVert < d_{\max}],
\end{equation}
where $d_{\max}\approx 1$ prevents the attractive force from collapsing distant objects. The cohesion term is a volume-weighted center-of-mass attractor with $\bar{\mathbf{p}}=\sum_i v_i\mathbf{p}_i / \sum_i v_i$, and centripetal pulls the cluster toward the container axis in the $xz$ plane:
\begin{equation}
\mathcal{L}_{\text{co}} = w_{\text{co}}\sum_i \lVert \mathbf{p}_i - \bar{\mathbf{p}}\rVert_2,\qquad
\mathcal{L}_{\text{cp}} = w_{\text{cp}}\sum_i \lVert \mathbf{p}_i^{xz} - \mathbf{c}^{xz}\rVert_2.
\end{equation}

Weights are $w_{\text{ov}}{=}80\ln(N{+}1)$, $w_{\text{bd}}{=}5.0$, $w_{\text{ct}}{=}2.0$, $w_{\text{co}}{=}0.5$, $w_{\text{cp}}{=}0.3$.
\begin{figure}[ht] 
\centering
\includegraphics[width=0.1\linewidth]{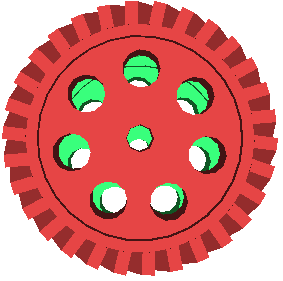}\hspace{0.2cm}
\includegraphics[width=0.1\linewidth]{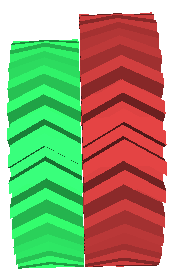}\hspace{0.2cm}
\includegraphics[width=0.1\linewidth]{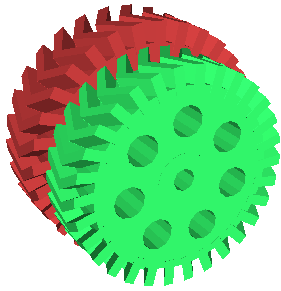}
\caption{Two gear objects are arranged in contact with each other(showing front, side, and back-top view), indicating tight packing with no overlaps. }
\label{fig:aabb_surrogate_gears}
\end{figure}

\subsection{Vectorized Pairwise Computation and Adaptive Squeezing}
\label{sec:vectorization}

Each loss term with an $i<j$ sum naively requires a double Python loop over $\binom{N}{2}$ pairs. We eliminate these loops by caching the AABBs as tensors $\mathbf{B}^-,\mathbf{B}^+\in\mathbb{R}^{N\times 3}$ and expressing every pairwise quantity through broadcasting:
\begin{equation}
\mathbf{O}_{ij,k} = \max(0, \min(B_{i,k}^+,B_{j,k}^+) - \max(B_{i,k}^-,B_{j,k}^-)),
\end{equation}
evaluated for all pairs in a single broadcast, producing an $N\times N\times 3$ tensor in one CUDA kernel launch. An upper-triangular mask selects unique pairs. The same pattern is applied to the contact gaps and the $xz$ footprints used by the gravity term. The speedup grows from 3.4$\times$ at $N{=}10$ to 54$\times$ at $N{=}100$.

\begin{figure}[!ht]
\centering
\includegraphics[width=0.95\linewidth]{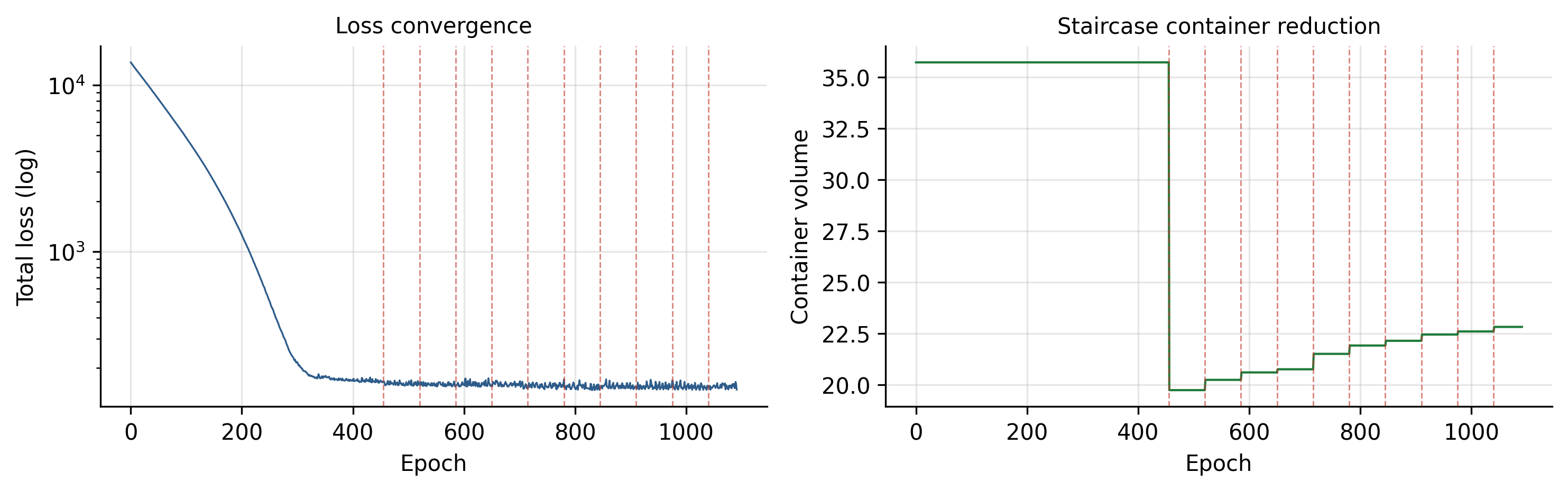}
\caption{Convergence dynamics on kitchen at $N{=}60$. Left: total loss (log scale). Right: container volume. Red dashed lines mark squeeze events. The first squeeze (epoch $\sim$450) delivers the large drop; later events stabilize the container at its equilibrium volume.}
\label{fig:convergence}
\end{figure}

\paragraph{Adaptive squeeze} Once overlaps are substantially resolved, the container is snapped to the tight AABB of all transformed vertices plus a margin:
\begin{equation}
\mathcal{C}_{\text{new}} = \mathrm{AABB}(\bigcup_i V_i') + \boldsymbol{\epsilon},\qquad
\epsilon = \max(0.08,\ 0.15/\sqrt[3]{N}).
\end{equation}
The $N^{-1/3}$ scaling follows from a dimensional argument: the characteristic single-object length scales as the cluster extent divided by $N^{1/3}$. A squeeze event fires only when the epoch is in $[0.10E_{\max},\,0.85E_{\max}]$, is a multiple of $T_s=\max(40,\lfloor E_{\max}/20\rfloor)$, and
\begin{equation}
\mathcal{L}_{\text{ov}} < \tau,\qquad \tau = \max(5.0,\ 0.005\cdot\tfrac{N(N-1)}{2}).
\end{equation}
The pair-count scaling of $\tau$ is key to operating reliably across $N$: a constant $\tau{=}5$ is strict at $N{=}10$ (45 pairs) and essentially always violated at $N{=}100$ (4{,}950 pairs). The scaled form converts $\tau$ into a bound on mean overlap per pair. Five to ten squeeze events typically occur during a single run. Figure~\ref{fig:convergence} illustrates the dynamics on kitchen at $N{=}60$. The loss drops three orders of magnitude in the first 300 epochs as gross overlaps are resolved. The first squeeze fires around epoch 450 once $\mathcal{L}_{\text{ov}} < \tau$, shrinking the container from roughly 35.5 to 19.7 in a single step. Subsequent squeezes make much smaller adjustments, and the container volume rises slightly between events as the optimizer trades off a tiny amount of tightness for the boundary and gravity terms it has to re-satisfy after each shrink. The final converged volume of around 22.8 reflects this equilibrium rather than continued monotone shrinking.

\subsection{Optimiser Configuration}
\label{sec:optconfig}

Two Adam~\cite{kingma2015adam} optimizers are used: lr$=0.005$ for translations, lr$=0.002$ for rotations (lower because a one-radian rotation moves far vertices by up to one unit while a unit translation moves every vertex by exactly one unit). Both use a \texttt{ReduceLROnPlateau} scheduler (patience 100, factor 0.7). Gradients are clipped to $L_2$ norm 0.5. Training runs for $E_{\max}=\min(5000,\,1000+5N)$ epochs with early stopping after $\min(500,\,200+N)$ epochs of no improvement. The initial container has volume $\sum_i v_i / 0.35$ with aspect ratio $(1.1, 0.85, 1.1)$. In fixed-container mode, the squeeze trigger is disabled, and the initial container is used throughout training; all other components are identical.

\FloatBarrier
\section{Experimental Setup}
\label{sec:experiments}

\subsection{Implementation and Datasets}

The framework uses Python 3 and PyTorch~\cite{paszke2019pytorch} 
for autodiff and GPU tensor operations, with 
\texttt{trimesh}~\cite{dawson2019trimesh} for mesh loading 
and \texttt{scipy.spatial.ConvexHull} for post-hoc hull 
volumes. Experiments run on a workstation with a 
consumer-grade GPU; each run uses a single GPU in 
\texttt{float32}. All runs are seeded.

We evaluate on multiple object categories:
\begin{itemize}
\item \textbf{Kitchen:} cups, bowls, plates, mugs, utensils; hollowness $h \approx 0.54$.
\item \textbf{Gear:} gear models with concave tooth profiles; $h \approx 0.40$.
\item \textbf{Block:} near-solid number-shaped blocks; $h \approx 0.88$.
\item \textbf{Mixed:} a union of the above; $h \approx 0.55$.
\end{itemize}
Hollowness $h = V_{\text{mesh}}/V_{\text{AABB}}$ is the ratio of mesh volume to bounding-box volume and matters when interpreting density: a thin-walled object cannot match a solid block's density even in principle. Each object is rescaled by its longest bounding-box edge to unit characteristic size, then recentered. Datasets are sampled with replacement and random scale in $[0.4, 0.8]$ to produce sets of size $N\in\{10,30,60,100\}$. We have also evaluated various other shapes, implying that our framework can be implemented on any 3D object.

\subsection{Evaluation Metrics}

\textbf{Container volume} $V = s_x s_y s_z$ is the headline metric (lower better). \textbf{Packing density} is $V_{\text{mesh}}/V_{\text{cont}}$, bounded above by $h$. \textbf{Packing efficiency} is the ratio of the sum of per-object convex-hull volumes to the global convex-hull volume, independent of container choice. \textbf{Container utilisation} is the ratio of the packed-cluster AABB to the container; values near $100\%$ indicate a tight squeeze, values above $100\%$ indicate overflow. \textbf{AABB overlap rate} is the fraction of pairs with intersection above a $2\%$-of-diagonal tolerance; this is an upper bound on true mesh overlap. Every reported result was inspected in MeshLab and has zero true mesh intersection. \textbf{Hull density} is the ratio of the sum of per-object AABB volumes to the convex-hull volume of the cluster, measuring how tightly the arrangement approximates its own bounding hull. \textbf{Floating count} is the number of objects whose lowest vertex sits more than $0.05$ unit-lengths above any supporting surface beneath them; it measures physical plausibility of the packing rather than its tightness.

\FloatBarrier
\section{Results I: Minimum-Container Packing (Squeeze Mode)}
\label{sec:odp_results}

We report our method in adaptive-squeeze mode on four categories at four object counts. Every entry in Table~\ref{tab:squeeze_results} is a mean over five seeds.

\paragraph{Container utilisation is consistent} Container utilisation stays in the $81$ to $95\%$ range across every dataset and every $N$. This is the signature of a tight squeeze: the container is pulled onto the converged cluster, rather than being left loose or forced to overflow. The same pair-count-scaled trigger handles $N{=}10$ and $N{=}100$ without per-instance tuning.

\paragraph{Block stands apart for geometric reasons} Block reaches $64$ to $71\%$ packing efficiency at $N\ge 30$; kitchen, gear, and mixed peak at $42$ to $55\%$. This reflects geometry, not the solver: block meshes are close to their AABBs ($h\approx 0.90$), so packing the AABB tightly also packs the mesh tightly. The solver-quality signal is two condition-independent metrics: hull density stays within $28$ to $66\%$ and container utilization within $81$ to $95\%$ across all categories.

\paragraph{Effect of $N$ within each dataset} Packing density behaves non-monotonically with $N$. On kitchen it decreases from $55\%$ at $N{=}10$ to $42\%$ at $N{=}60$ before rising back to $47\%$ at $N{=}100$. The dip at intermediate $N$ reflects a geometric transition: at $N{=}10$ the cluster is essentially a single layer and the squeeze is bounded by the largest object's footprint; at $N{=}60$ the cluster is tall enough that vertical voids open between stacking layers; by $N{=}100$ enough small objects exist to fill those voids, and density recovers. Block and gear follow the same pattern with different inflection points dictated by their hollowness factors. This non-monotonic behavior is a property of the geometry, not the solver: container utilization stays in the $81$ to $95\%$ range at every $N$, confirming the squeeze mechanism converges equally tightly regardless of which phase the cluster is in.

\begin{figure*}[!htbp]
\centering
\begin{minipage}{0.04\textwidth}\centering{\scriptsize\rotatebox{90}{\textbf{Kitchen}}}\end{minipage}%
\begin{subfigure}[b]{0.19\textwidth}\imgOrPlaceholder{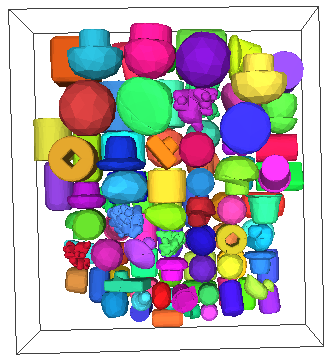}{\textwidth}\end{subfigure}\hfill
\begin{subfigure}[b]{0.19\textwidth}\imgOrPlaceholder{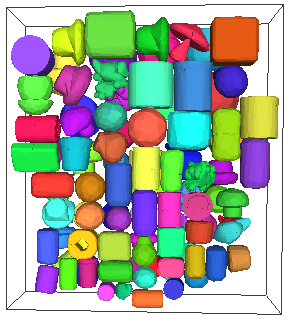}{\textwidth}\end{subfigure}\hfill
\begin{subfigure}[b]{0.19\textwidth}\imgOrPlaceholder{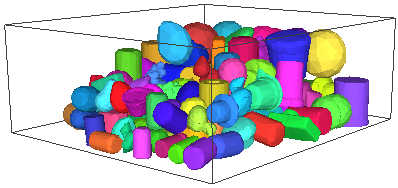}{\textwidth}\end{subfigure}\hfill
\begin{subfigure}[b]{0.19\textwidth}\imgOrPlaceholder{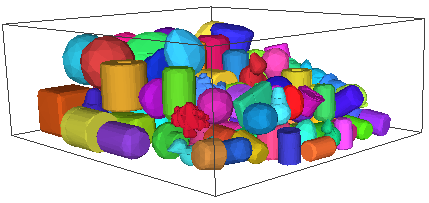}{\textwidth}\end{subfigure}

\vspace{1pt}
\begin{minipage}{0.04\textwidth}\centering{\scriptsize\rotatebox{90}{\textbf{Gear}}}\end{minipage}%
\begin{subfigure}[b]{0.22\textwidth}\imgOrPlaceholder{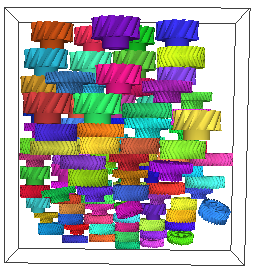}{\textwidth}\end{subfigure}\hfill
\begin{subfigure}[b]{0.22\textwidth}\imgOrPlaceholder{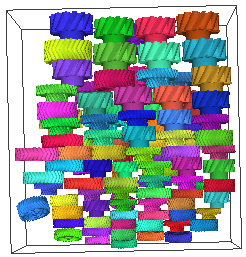}{\textwidth}\end{subfigure}\hfill
\begin{subfigure}[b]{0.22\textwidth}\imgOrPlaceholder{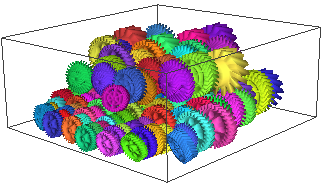}{\textwidth}\end{subfigure}\hfill
\begin{subfigure}[b]{0.22\textwidth}\imgOrPlaceholder{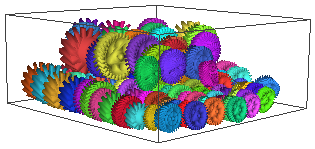}{\textwidth}\end{subfigure}

\vspace{1pt}
\begin{minipage}{0.04\textwidth}\centering{\scriptsize\rotatebox{90}{\textbf{Block}}}\end{minipage}%
\begin{subfigure}[b]{0.22\textwidth}\imgOrPlaceholder{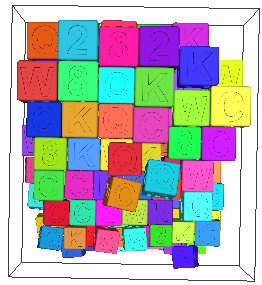}{\textwidth}\end{subfigure}\hfill
\begin{subfigure}[b]{0.22\textwidth}\imgOrPlaceholder{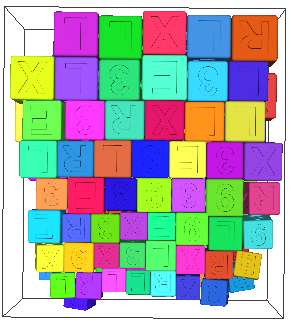}{\textwidth}\end{subfigure}\hfill
\begin{subfigure}[b]{0.22\textwidth}\imgOrPlaceholder{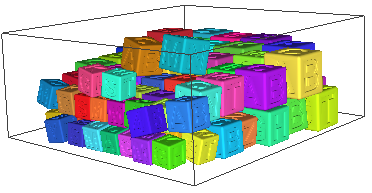}{\textwidth}\end{subfigure}\hfill
\begin{subfigure}[b]{0.22\textwidth}\imgOrPlaceholder{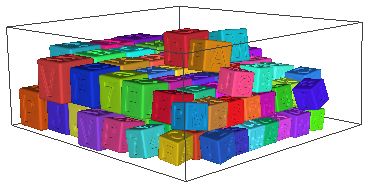}{\textwidth}\end{subfigure}

\vspace{1pt}
\begin{minipage}{0.04\textwidth}\centering{\scriptsize\rotatebox{90}{\textbf{Mixed}}}\end{minipage}%
\begin{subfigure}[b]{0.22\textwidth}\imgOrPlaceholder{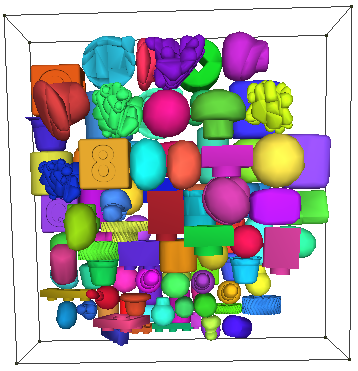}{\textwidth}\end{subfigure}\hfill
\begin{subfigure}[b]{0.22\textwidth}\imgOrPlaceholder{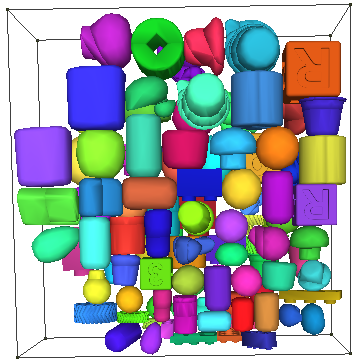}{\textwidth}\end{subfigure}\hfill
\begin{subfigure}[b]{0.22\textwidth}\imgOrPlaceholder{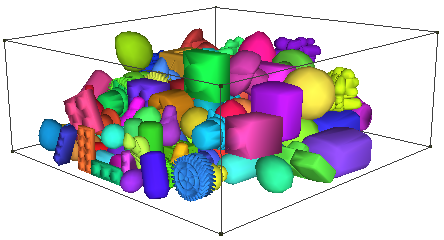}{\textwidth}\end{subfigure}\hfill
\begin{subfigure}[b]{0.22\textwidth}\imgOrPlaceholder{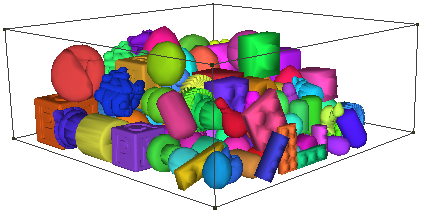}{\textwidth}\end{subfigure}

\caption{Squeeze-mode results at $N{=}100$. Four views (top, bottom, right, left) per dataset row. All configurations have zero true mesh-mesh overlap.}
\label{fig:odp_grid}
\end{figure*}

\paragraph{Overlap and runtime} AABB overlap rate stays at or below $0.4\%$ on every cell, and $0.0\%$ on blocks at $N\le 30$. Every arrangement has zero true mesh intersection (Figure~\ref{fig:odp_grid}). Residual AABB overlaps are concave objects whose bounding boxes touch without the meshes intersecting. Runtime scales roughly linearly with $N$: from $24{-}37$s at $N{=}10$ to $218{-}223$s at $N{=}100$.

\begin{table*}[!htbp]
\centering
\caption{Minimum-container (squeeze) packing results. All configurations have zero true mesh-mesh intersections, verified in MeshLab.}
\label{tab:squeeze_results}
\resizebox{\textwidth}{!}{%
\begin{tabular}{llccccccccc}
\toprule
Dataset & $n$ & Container & Pack Eff.\% & Hull Dens.\% & Cont.\ Util.\% & Overall\% & $h$ & AABB Ov. & Mesh Ov. & Time \\
\midrule
\multirow{4}{*}{Kitchen}
  & 10  & $1.604 \times 0.876 \times 1.961$ & 55.0 & 51.9& 82.3& 32.4 & 0.56 & 2.2& 0 & 36.2s \\
  & 30  & $2.761 \times 1.498 \times 2.602$ & 45.7& 42.8 & 89.1 & 24.0& 0.54 & 0.2 & 0 & 69.7s \\
  & 60  & $3.739 \times 1.725 \times 3.860$ &42.3 & 38.7 & 91.7 & 26.3 & 0.53 & 0.1 & 0 & 106.7s \\
  & 100 & $4.126 \times 1.681 \times 4.583$ & 47.3 & 43.6 & 94.8 & 27.4 & 0.54 & 0.1 & 0 & 218.9s \\
\midrule
\multirow{4}{*}{Gear}
  & 10  & $1.522 \times 0.866 \times 1.511$ & 63.1 & 38.5 & 81.3 & 26.9 & 0.43 & 0   & 0 & 24.3s \\
  & 30  & $2.784 \times 1.585 \times 2.424$ & 47.3 & 28.0 & 88.7 & 14.9 & 0.41 & 0.2 & 0 & 58.0s \\
  & 60  & $3.287 \times 1.629 \times 3.376$ & 48.7 & 28.5 & 90.6 & 15.9 & 0.40 & 0.3 & 0 & 132.8s \\
  & 100 & $3.675 \times 1.583 \times 3.877$ & 51.2 & 29.8 & 89.4 & 20.2 & 0.40 & 0.1 & 0 & 220.8s \\
\midrule
\multirow{4}{*}{Block}
  & 10  & $1.935 \times 1.522 \times 2.609$ & 64.5 & 60.6 & 88.1 & 38.6 & 0.88 & 0   & 0 & 32.9s \\
  & 30  & $3.039 \times 1.616 \times 3.349$ & 70.9 & 66.4 & 90.7 & 51.5 & 0.87 & 0   & 0 & 64.8s \\
  & 60  & $4.249 \times 1.859 \times 4.478$ & 67.6 & 63.3 & 91.0 & 42.3 & 0.88 & 0.2 & 0 & 133.9s \\
  & 100 & $4.695 \times 2.012 \times 5.174$ & 67.7 & 63.3 & 94.6 & 45.7 & 0.86 & 0.1 & 0 & 223.1s \\
\midrule
\multirow{4}{*}{Mixed}
  & 10  & $2.028 \times 1.387 \times 1.655$ & 55.0 & 47.6 & 85.8 & 26.3 & 0.63 & 0   & 0 & 35.3s \\
  & 30  & $2.883 \times 1.484 \times 2.658$ & 50.2 & 42.3 & 89.3 & 27.3 & 0.51 & 0.2 & 0 & 70.4s \\
  & 60  & $3.764 \times 1.589 \times 3.837$ & 47.0 & 40.2 & 92.2 & 24.5 & 0.52 & 0.3 & 0 & 132.7s \\
  & 100 & $4.369 \times 1.753 \times 4.472$ & 45.3 & 39.8 & 92.5 & 25.3 & 0.54 & 0.2 & 0 & 221.0s \\
\bottomrule
\end{tabular}}
\end{table*}
\FloatBarrier
\section{Results II: Fixed-Container Packing (No-Squeeze Mode)}
\label{sec:bpp_results}

\begin{figure*}[!htbp]
\centering
\begin{subfigure}[b]{0.19\textwidth}\imgOrPlaceholder{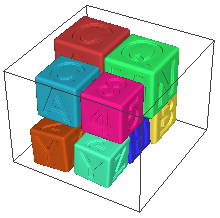}{\textwidth}\caption{Block $n{=}10$}\end{subfigure}\hfill
\begin{subfigure}[b]{0.19\textwidth}\imgOrPlaceholder{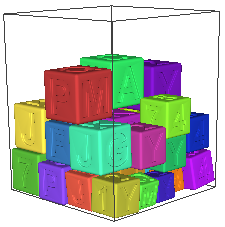}{\textwidth}\caption{Block $n{=}30$}\end{subfigure}\hfill
\begin{subfigure}[b]{0.19\textwidth}\imgOrPlaceholder{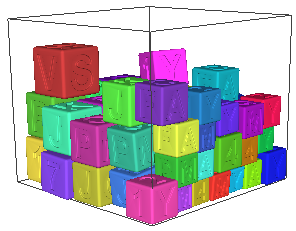}{\textwidth}\caption{Block $n{=}60$}\end{subfigure}\hfill
\begin{subfigure}[b]{0.19\textwidth}\imgOrPlaceholder{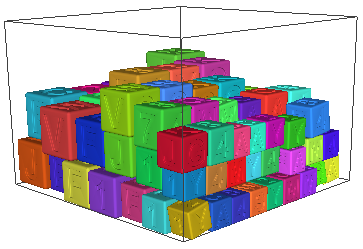}{\textwidth}\caption{Block $n{=}100$}\end{subfigure}

\vspace{2pt}
\begin{subfigure}[b]{0.19\textwidth}\imgOrPlaceholder{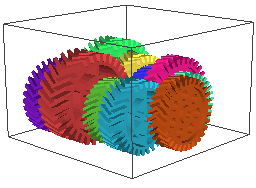}{\textwidth}\caption{Gear $n{=}10$}\end{subfigure}\hfill
\begin{subfigure}[b]{0.19\textwidth}\imgOrPlaceholder{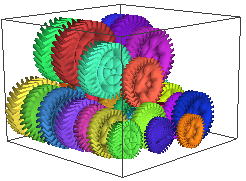}{\textwidth}\caption{Gear $n{=}30$}\end{subfigure}\hfill
\begin{subfigure}[b]{0.19\textwidth}\imgOrPlaceholder{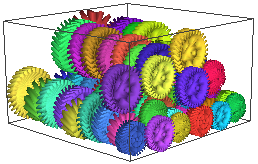}{\textwidth}\caption{Gear $n{=}60$}\end{subfigure}\hfill
\begin{subfigure}[b]{0.19\textwidth}\imgOrPlaceholder{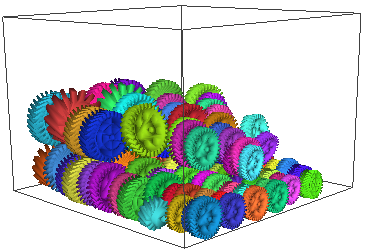}{\textwidth}\caption{Gear $n{=}100$}\end{subfigure}

\vspace{2pt}
\begin{subfigure}[b]{0.19\textwidth}\imgOrPlaceholder{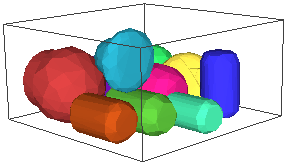}{\textwidth}\caption{Kitchen $n{=}10$}\end{subfigure}\hfill
\begin{subfigure}[b]{0.19\textwidth}\imgOrPlaceholder{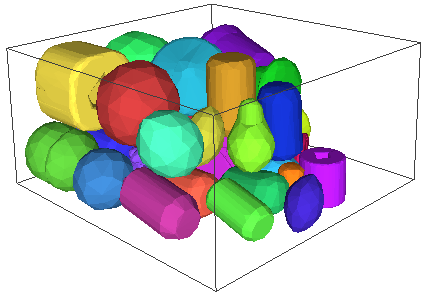}{\textwidth}\caption{Kitchen $n{=}30$}\end{subfigure}\hfill
\begin{subfigure}[b]{0.19\textwidth}\imgOrPlaceholder{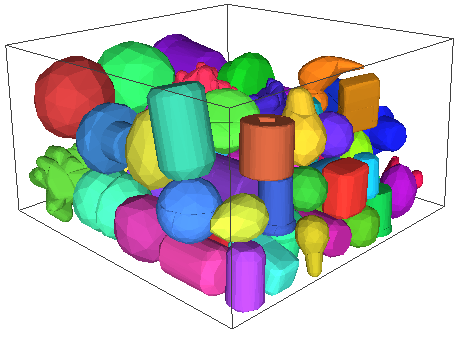}{\textwidth}\caption{Kitchen $n{=}60$}\end{subfigure}\hfill
\begin{subfigure}[b]{0.19\textwidth}\imgOrPlaceholder{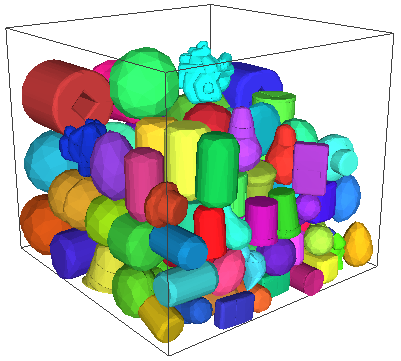}{\textwidth}\caption{Kitchen $n{=}100$}\end{subfigure}

\caption{Fixed-container (no-squeeze) packings. Mixed-scale objects with random scale in $[0.4, 0.8]$. All configurations have zero mesh-mesh overlap despite AABB overlaps reported in Table~\ref{tab:bpp}. Note the emergent multi-layer stacking, particularly visible in blocks at small $n$.}
\label{fig:bpp_grid}
\end{figure*}

In fixed-container mode the squeezing mechanism is disabled and container dimensions are specified as input. We evaluate on three categories at four $N$ values. Container dimensions are chosen to provide a feasible but challenging packing problem. To reflect real-world variability, objects within each category are assigned random scale factors in $[0.4, 0.8]$, in contrast to prior work~\cite{zhuang2024dynamics,ma2018packing} where instances share identical dimensions. Table~\ref{tab:bpp} reports all metrics; Figure~\ref{fig:bpp_grid} visualises the arrangements.

\begin{table*}[!htbp]
\centering
\caption{Fixed-container (no-squeeze) packing results. ``AABB Ov.'' reports axis-aligned bounding-box overlaps arising from proxy conservatism on concave objects; all configurations have zero true mesh-mesh intersections.}
\label{tab:bpp}
\resizebox{\textwidth}{!}{%
\begin{tabular}{llccccccccc}
\toprule
Dataset & $n$ & Container & Pack Eff.\% & Hull Dens.\% & Cont.\ Util.\% & Overall\% & $h$ & AABB Ov. & Mesh Ov. & Time \\
\midrule
\multirow{4}{*}{Block}
  & 10  & $2.0 \times 1.5 \times 2.0$   & 79.8 & 74.4 & 73.7  & 43.2 & 0.87 & 0  & 0 & 37s \\
  & 30  & $2.5 \times 3.0 \times 2.5$   & 71.0 & 66.4 & 77.6  & 40.9 & 0.89 & 0  & 0 & 50s \\
  & 60  & $3.3 \times 2.8 \times 3.3$   & 66.7 & 62.4 & 90.2  & 44.6 & 0.88 & 2  & 0 & 131s \\
  & 100 & $4.1 \times 2.8 \times 4.1$   & 74.3 & 69.6 & 80.3  & 44.9 & 0.90 & 5  & 0 & 184s \\
\midrule
\multirow{4}{*}{Gear}
  & 10  & $1.5 \times 1.0 \times 1.5$   & 69.9 & 35.5 & 69.0  & 17.9 & 0.38 & 0  & 0 & 28s \\
  & 30  & $2.0 \times 1.5 \times 2.0$   & 60.2 & 33.6 & 98.6  & 20.4 & 0.41 & 16 & 0 & 42s \\
  & 60  & $2.5 \times 1.5 \times 2.5$   & 55.9 & 31.7 & 107.9 & 24.6 & 0.41 & 16 & 0 & 129s \\
  & 100 & $3.5 \times 2.5 \times 3.5$   & 50.7 & 29.2 & 58.1  & 11.8 & 0.40 & 7  & 0 & 106s \\
\midrule
\multirow{4}{*}{Kitchen}
  & 10  & $2.0 \times 1.0 \times 2.0$   & 52.7 & 49.0 & 94.2  & 26.7 & 0.58 & 0  & 0 & 20s \\
  & 30  & $2.8 \times 1.4 \times 2.8$   & 51.2 & 47.0 & 89.9  & 29.1 & 0.55 & 0  & 0 & 43s \\
  & 60  & $3.0 \times 1.7 \times 3.0$   & 48.8 & 45.1 & 107.9 & 36.4 & 0.55 & 4  & 0 & 107s \\
  & 100 & $3.0 \times 2.7 \times 3.0$   & 49.9 & 46.1 & 104.8 & 36.4 & 0.56 & 26 & 0 & 215s \\
\bottomrule
\end{tabular}}
\end{table*}

\textbf{Blocks achieve the highest packing efficiency} (74--80\% at $n\le 30$) with zero AABB overlaps, owing to their near-solid geometry; the AABB proxy closely approximates the actual mesh. \textbf{Gears exhibit AABB overlaps that are not mesh overlaps}: at $n{=}30$ and $n{=}60$, 16 AABB-overlapping pairs are reported despite zero actual mesh intersection, because interlocking teeth create overlapping bounding boxes even when the meshes maintain clearance. \textbf{Emergent multi-layer stacking} appears across all three datasets without explicit heuristics, driven by the interaction of support-aware gravity, contact attraction, and cohesion. \textbf{Container utilization above $100\%$} (gear $n{=}60$: 107.9\%, kitchen $n{=}60$: 107.9\%, kitchen $n{=}100$: 104.8\%) indicates the packed AABB extends slightly beyond the container walls at contact points; visual inspection confirms all objects remain tangent to or inside the container.

\FloatBarrier
\section{Comparison with Time-Matched Baselines}
\label{sec:results_baselines}

We compare against three standard baselines on three datasets at four object counts. Protocol: identical inputs (same meshes, same five seeds), wall-clock budget for metaheuristic baselines matched to the median runtime of our method.

\paragraph{Baselines} \textbf{DBLF}~\cite{wang2010two} places objects sequentially at the deepest, bottom-most, left-most extreme point admitting zero overlap, implemented with the full extreme-point set of Crainic et al.~\cite{crainic2008extreme} and 24 canonical orientations per object. \textbf{SA} runs simulated annealing on both translations and rotations from random initialisation, with adaptive step size and an overlap-volume penalty. \textbf{BLF+SA} initialises from DBLF and refines with SA. We use the same vertex transformation and AABB evaluation as our method, so all four reported volumes are computed identically.

\begin{table*}[!htbp]
\centering
\caption{Final container volume (lower is better). Mean and standard deviation over 5 seeds. Bold: best per row.}
\label{tab:baseline_compare}
\small
\begin{tabular}{llcccc}
\toprule
Dataset & $N$ & DBLF & SA & BLF+SA & \textbf{Ours} \\
\midrule
\multirow{4}{*}{Block}
 & 10  & $\mathbf{4.59 \pm 1.05}$ & $7.60 \pm 2.46$ & $4.59 \pm 1.05$ & $5.19 \pm 2.06$ \\
 & 30  & $16.39 \pm 1.39$ & $26.78 \pm 4.61$ & $16.39 \pm 1.39$ & $\mathbf{16.18 \pm 3.32}$ \\
 & 60  & $35.08 \pm 1.88$ & $53.49 \pm 5.45$ & $35.08 \pm 1.88$ & $\mathbf{33.49 \pm 0.99}$ \\
 & 100 & $60.34 \pm 2.20$ & $90.68 \pm 4.78$ & $60.34 \pm 2.20$ & $\mathbf{53.96 \pm 2.36}$ \\
\midrule
\multirow{4}{*}{Gear}
 & 10  & $2.24 \pm 0.59$ & $2.15 \pm 0.48$ & $1.80 \pm 0.35$ & $\mathbf{1.63 \pm 0.48}$ \\
 & 30  & $8.20 \pm 1.09$ & $10.35 \pm 1.80$ & $8.20 \pm 1.09$ & $\mathbf{8.01 \pm 1.18}$ \\
 & 60  & $18.44 \pm 0.76$ & $25.35 \pm 3.70$ & $18.44 \pm 0.75$ & $\mathbf{16.07 \pm 1.35}$ \\
 & 100 & $29.60 \pm 2.54$ & $42.86 \pm 4.15$ & $29.60 \pm 2.54$ & $\mathbf{24.12 \pm 1.54}$ \\
\midrule
\multirow{4}{*}{Kitchen}
 & 10  & $3.47 \pm 0.89$ & $\mathbf{2.35 \pm 0.55}$ & $2.44 \pm 0.67$ & $3.58 \pm 1.23$ \\
 & 30  & $13.44 \pm 1.75$ & $12.33 \pm 1.50$ & $12.98 \pm 1.60$ & $\mathbf{12.03 \pm 2.36}$ \\
 & 60  & $28.37 \pm 1.48$ & $36.36 \pm 3.60$ & $28.37 \pm 1.48$ & $\mathbf{22.16 \pm 2.69}$ \\
 & 100 & $49.38 \pm 2.77$ & $66.63 \pm 6.19$ & $49.38 \pm 2.77$ & $\mathbf{36.52 \pm 1.46}$ \\
\bottomrule
\end{tabular}
\end{table*}

\paragraph{Result and observations} Our method produces the smallest container at every $(N, \text{dataset})$ pair for $N\ge 30$, with reductions of $11{-}22\%$ over DBLF and $32{-}45\%$ over time-matched SA at $N{=}100$. At $N{=}10$, the small search space favors the metaheuristic baselines on two of three datasets. Three observations: (1) DBLF and BLF+SA produce identical numbers at $N\ge 30$, meaning the SA refinement layer found zero improving moves on top of DBLF's construction within the time budget; (2) time-matched SA from random initialisation degrades with $N$ on block and gear, returning containers \emph{larger} than the initial estimate at $N\ge 60$; (3) our advantage grows with $N$, from a tie at $N{=}30$ on block to $13\%$ reduction at $N{=}100$, and from $1\%$ at $N{=}30$ on kitchen to $26\%$ at $N{=}100$.

\paragraph{Baseline asymmetries} The baselines use a more constrained rotation search than our method's continuous gradient optimization; some of our advantage at large 
$N$ is therefore attributable to the additional effective degrees of freedom rather than to the squeeze mechanism alone.

\paragraph{Note on overlap} Constructive baselines produce zero AABB overlap by construction. Our method's residual AABB overlap is non-zero (typically $3{-}5\%$ of pairs), but every result has zero true mesh intersection: the AABBs of two concave objects can touch while the meshes do not. A tighter overlap surrogate is discussed in Section~\ref{sec:limitations}.

\begin{figure*}[!htbp]
\centering
\begin{minipage}{0.05\textwidth}\centering{\scriptsize\rotatebox{90}{\textbf{Block}}}\end{minipage}%
\begin{subfigure}[b]{0.20\textwidth}\imgOrPlaceholder{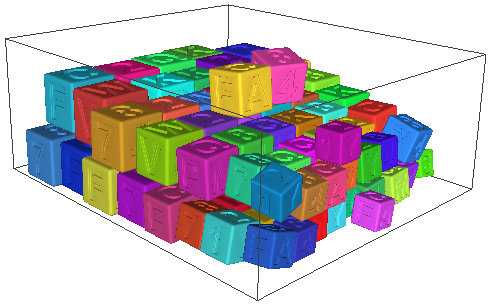}{\textwidth}\caption*{\scriptsize Ours (53.96)}\end{subfigure}\hfill
\begin{subfigure}[b]{0.20\textwidth}\imgOrPlaceholder{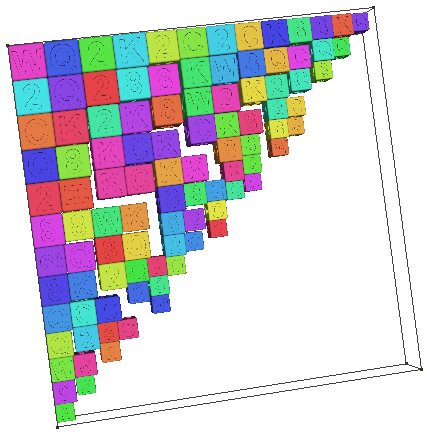}{\textwidth}\caption*{\scriptsize DBLF (60.34)}\end{subfigure}\hfill
\begin{subfigure}[b]{0.20\textwidth}\imgOrPlaceholder{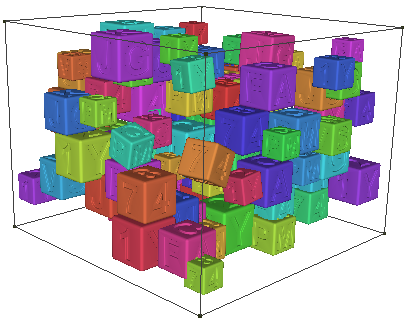}{\textwidth}\caption*{\scriptsize SA (90.68)}\end{subfigure}\hfill
\begin{subfigure}[b]{0.20\textwidth}\imgOrPlaceholder{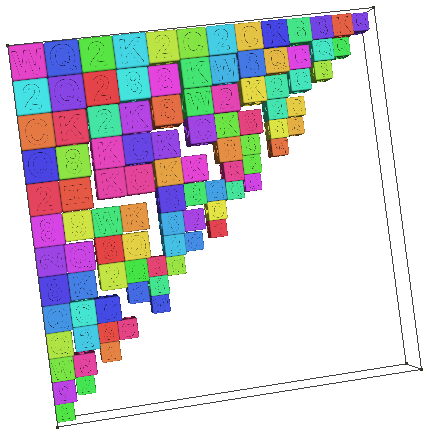}{\textwidth}\caption*{\scriptsize BLF+SA (60.34)}\end{subfigure}

\vspace{2pt}
\begin{minipage}{0.05\textwidth}\centering{\scriptsize\rotatebox{90}{\textbf{Gear}}}\end{minipage}%
\begin{subfigure}[b]{0.20\textwidth}\imgOrPlaceholder{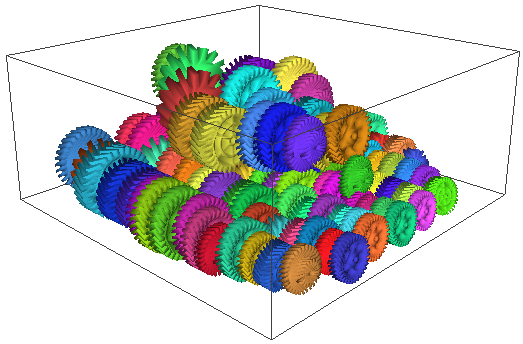}{\textwidth}\caption*{\scriptsize Ours (24.12)}\end{subfigure}\hfill
\begin{subfigure}[b]{0.20\textwidth}\imgOrPlaceholder{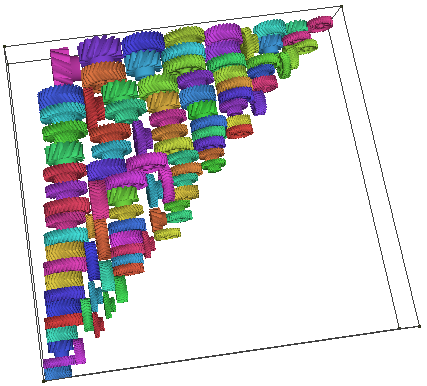}{\textwidth}\caption*{\scriptsize DBLF (29.60)}\end{subfigure}\hfill
\begin{subfigure}[b]{0.20\textwidth}\imgOrPlaceholder{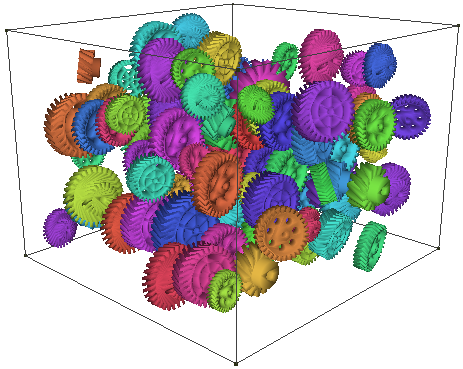}{\textwidth}\caption*{\scriptsize SA (42.86)}\end{subfigure}\hfill
\begin{subfigure}[b]{0.20\textwidth}\imgOrPlaceholder{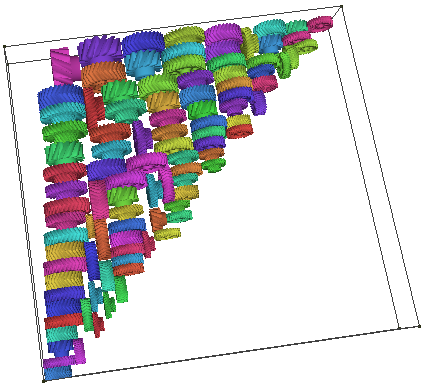}{\textwidth}\caption*{\scriptsize BLF+SA (29.60)}\end{subfigure}

\vspace{2pt}
\begin{minipage}{0.05\textwidth}\centering{\scriptsize\rotatebox{90}{\textbf{Kitchen}}}\end{minipage}%
\begin{subfigure}[b]{0.20\textwidth}\imgOrPlaceholder{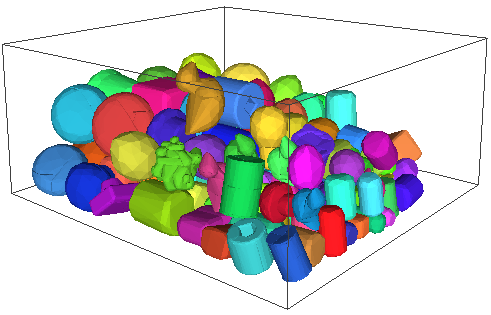}{\textwidth}\caption*{\scriptsize Ours (36.52)}\end{subfigure}\hfill
\begin{subfigure}[b]{0.20\textwidth}\imgOrPlaceholder{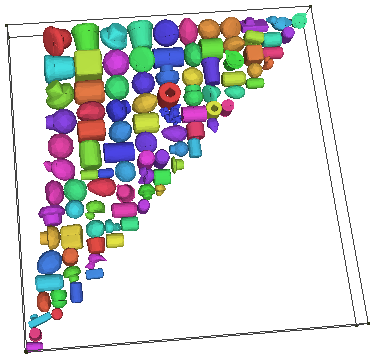}{\textwidth}\caption*{\scriptsize DBLF (49.38)}\end{subfigure}\hfill
\begin{subfigure}[b]{0.20\textwidth}\imgOrPlaceholder{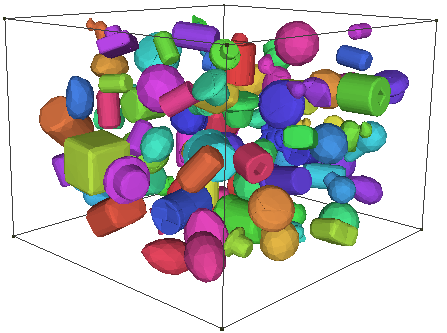}{\textwidth}\caption*{\scriptsize SA (66.63)}\end{subfigure}\hfill
\begin{subfigure}[b]{0.20\textwidth}\imgOrPlaceholder{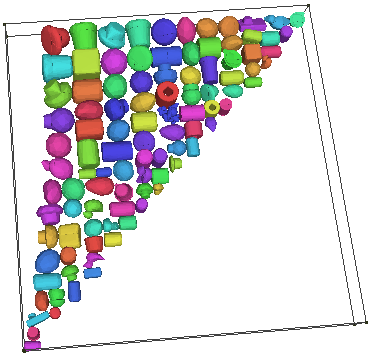}{\textwidth}\caption*{\scriptsize BLF+SA (49.38)}\end{subfigure}

\caption{Visual comparison at $N{=}100$, seed $0$. Container volumes in parentheses. Our method (left column) yields the most compact arrangement on every dataset.}
\label{fig:baseline_grid}
\end{figure*}
\begin{table}[!htbp]
\centering
\caption{Mean wall-clock time (s) per run, averaged across block, gear, and kitchen at each $N$. SA and BLF+SA are budget-matched to Ours.}
\label{tab:timing}
\small
\begin{tabular}{lcccc}
\toprule
Method & $N{=}10$ & $N{=}30$ & $N{=}60$ & $N{=}100$ \\
\midrule
DBLF     & 0.3 & 6.0   & 44.8  & 202.3 \\
SA       & 41.0 & 81.9 & 140.9 & 231.2 \\
BLF+SA   & 41.2 & 87.9 & 185.8 & 433.8 \\
\textbf{Ours} & \textbf{40.9} & \textbf{81.3} & \textbf{131.1} & \textbf{239.8} \\
\bottomrule
\end{tabular}
\end{table}

\FloatBarrier
\section{Ablation Study}
\label{sec:ablation}

We disable one component at a time to isolate what each contributes. All runs use the kitchen dataset at $N{=}60$ with five random seeds. Three metrics are reported: container volume (tightness), overlap count (feasibility), and floating count (objects whose bottom sits more than $0.05$ units above any supporting surface; physical plausibility). A well-factorized loss decomposition should have the property that each ablation damages a \emph{different} metric.

\begin{figure}[H] %
\centering

\captionof{table}{Ablation at $N{=}60$. Mean $\pm$ std, 5 seeds.}
\label{tab:ablation}
\small
\begin{tabular}{lccc}
\toprule
Variant & Vol & Ov. & Float \\
\midrule
Full          & $19.12{\pm}1.67$ & $2.2{\pm}2.5$ & $\mathbf{0.2{\pm}0.4}$ \\
No squeeze    & $\mathbf{18.32{\pm}1.29}$ & $4.0{\pm}1.7$ & $1.4{\pm}0.5$ \\
No gravity    & $20.74{\pm}2.59$ & $\mathbf{0.6{\pm}0.5}$ & $23.8{\pm}1.2$ \\
No coh./cent. & $19.95{\pm}1.45$ & $1.4{\pm}1.0$ & $1.0{\pm}1.3$ \\
No rotation   & $19.20{\pm}1.94$ & $6.4{\pm}2.2$ & $1.4{\pm}1.0$ \\
\bottomrule
\end{tabular}

\vspace{1em} 


\includegraphics[width=0.85\linewidth, trim={0cm 0cm 0cm 0cm}, clip]{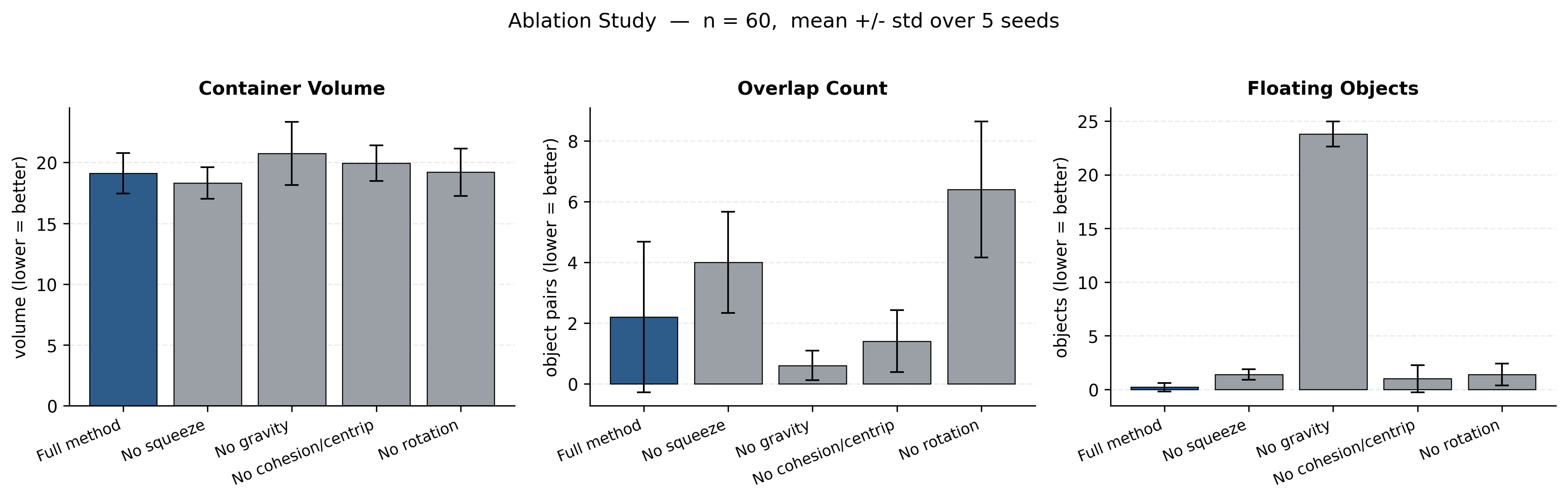}
\captionof{figure}{Ablation metrics; full method (blue) vs ablations (grey).}
\label{fig:ablation_plot}

\end{figure}
\paragraph{Reading the variants} \emph{No gravity} removes the vertical pull toward supporting surfaces. The floating count rises from $0.2$ to $23.8$ out of $60$ objects, while container volume and overlap remain essentially unchanged. \emph{No rotation} constrains the optimizer to three translational degrees of freedom; container volume is barely affected, but overlap count rises from $2.2$ to $6.4$ pairs, since the optimizer cannot rotate objects into interlocking configurations. \emph{No cohesion and centripetal} removes the soft regularisers; container volume rises from $19.12$ to $19.95$ ($+4\%$) because the contact term alone cannot pull a fragmented cluster back together.

\emph{No squeeze} is more subtle. The reported $18.32$ is numerically smaller than the full method's $19.12$, but this does not mean squeezing hurts. Without squeeze the container remains at the initial estimate throughout training, and the reported volume is the tight AABB of the converged cluster inside that oversized envelope. The full method pulls the container down during optimization, and the cluster at convergence is denser and vertically taller, yielding a slightly larger tight AABB. The consequences of disabling squeeze are captured in container utilization (Section~\ref{sec:odp_results}): with squeeze disabled, the cluster either leaves substantial slack or overflows, whereas the full method consistently achieves $81$ to $95\%$ utilization.

\paragraph{Summary} Every component addresses a distinct failure: gravity for stability, rotation for feasibility, cohesion/centripetal for volume tightening, and squeeze for container sizing. Each ablation breaks a different metric, confirming a well-factorized design.

\FloatBarrier
\section{Limitations}
\label{sec:limitations}

\paragraph{AABB overlap surrogate} Overlap is computed between axis-aligned bounding boxes, not between meshes. For irregular meshes, this is a strict upper bound on true mesh overlap, so the optimizer can converge with small residual AABB overlap while meshes do not intersect. A tighter surrogate based on oriented bounding boxes or a differentiable signed distance field \cite{park2019deepsdf} would further reduce container size at the cost of additional per-epoch computation.

\paragraph{Small $N$} At $N\le 10$, simulated annealing outperforms our method on some datasets. The gradient-based approach has its advantage at scale ($N\ge 60$), where the high-dimensional configuration space overwhelms random perturbation but the gradient still carries useful information.

\paragraph{Published-method comparison} Direct head-to-head comparison with recent developments was not possible because the specific 3D models used in their experiments are not released as a public dataset. We therefore compare only against baselines we can re-implement on our datasets. Context from their published numbers is available in the respective papers.

\paragraph{Rectangular container} The formulation assumes a right rectangular prism container. Extending the boundary loss to general convex containers defined by a signed distance function is straightforward in principle but was not pursued here.

\paragraph{Off-line setting} The method is an off-line, all-at-once packer. It does not address the online variant, for which learning-based approaches retain a structural advantage.

\paragraph{Future work} Promising extensions: GPU-level parallelization for several hundred objects; differentiable SDF or OBB overlap surrogates; non-rectangular containers; and a light-weight learned initializer that warm-starts the optimizer.

\FloatBarrier
\section{Conclusion}
\label{sec:conclusion}

We presented a differentiable optimization framework for packing irregular 3D objects that jointly estimates all three container dimensions. The formulation combines six physics-inspired loss terms with an adaptive container squeezing mechanism whose pair-count-scaled threshold makes the same pipeline work reliably from $N{=}10$ to $N{>}100$ without per-instance tuning. Vectorizing every pairwise computation yields a $3.4$-$ 54$-fold speedup over the reference Python loop, and the full pipeline finishes a 100-object instance in under 4 minutes on a single GPU using only Python and PyTorch. Across four object categories, the method produces overlap-free arrangements with containers $11$ to $32\%$ smaller than time-matched DBLF and SA baselines at $N\ge 60$. The Python-only design removes the C\texttt{++}, CUDA, and PhysX dependency of the strongest existing pipelines, which we expect to make the method substantially easier to integrate into larger differentiable systems.

\section*{Acknowledgements}
The first author gratefully acknowledges the support of the 
India AI Fellowship under the IndiaAI Mission.

\FloatBarrier
\bibliographystyle{elsarticle-num}

\end{document}